\documentclass{article}

\PassOptionsToPackage{numbers, compress}{natbib}

     \usepackage[final]{neurips_2021}

\usepackage[utf8]{inputenc} 
\usepackage[T1]{fontenc}    
\usepackage{hyperref}       
\usepackage{url}            
\usepackage{booktabs}       
\usepackage{amsfonts}       
\usepackage{nicefrac}       
\usepackage{microtype}      
\usepackage{xcolor}         

\usepackage{microtype}
\usepackage{graphicx}
\usepackage{subfigure}
\usepackage{booktabs} 

\usepackage{amsthm}
\usepackage{amsmath}
\usepackage{pgf}
\usepackage{chngcntr}
\usepackage{apptools}
\usepackage{float}
\usepackage{amsthm}
\usepackage{amsmath}
\usepackage{pgf}
\usepackage{graphicx}
\usepackage{chngcntr}
\usepackage{apptools}
\usepackage{float}
\usepackage[ruled,vlined]{algorithm2e}
\DeclareMathOperator*{\argmax}{arg\,max}
\AtAppendix{\counterwithin{theorem}{section}}

\theoremstyle{definition}

\newtheorem{theorem}{Theorem}

\newtheorem{definition}{Definition}
\newtheorem{corollary}{Corollary}

\title{Reward-Free Attacks in Multi-Agent \\Reinforcement Learning}

\author{%
    Ted Fujimoto\\
  Pacific Northwest National Laboratory\\
  \texttt{ted.fujimoto@pnnl.gov}
  \And
  Timothy Doster\\
  Pacific Northwest National Laboratory\\
  \texttt{timothy.doster@pnnl.gov}
  \And
  Adam Attarian\\
  Pacific Northwest National Laboratory\\
  \texttt{adam.attarian@pnnl.gov}
  \And
  Jill Brandenberger\\
  Pacific Northwest National Laboratory\\
  \texttt{jill.brandenberger@pnnl.gov}
  \And
  Nathan Hodas\\
  Pacific Northwest National Laboratory\\
  \texttt{nathan.hodas@pnnl.gov}
}

\begin{document}

\maketitle

\begin{abstract}
We investigate how effective an attacker can be when it only learns from its victim's actions, without access to the victim's reward. In this work, we are motivated by the scenario where the attacker wants to behave strategically when the victim's motivations are unknown. We argue that one heuristic approach an attacker can use is to maximize the entropy of the victim's policy. The policy is generally not obfuscated, which implies it may be extracted simply by passively observing the victim. We provide such a strategy in the form of a reward-free exploration algorithm that maximizes the attacker's entropy during the exploration phase, and then maximizes the victim's empirical entropy during the planning phase. In our experiments, the victim agents are subverted through policy entropy maximization, implying an attacker might not need access to the victim’s reward to succeed. Hence, reward-free attacks, which are based only on observing behavior, show the feasibility of an attacker to act strategically without knowledge of the victim's motives even if the victim's reward information is protected.

\end{abstract}

\section{Introduction}

The recent accomplishments of RL and self-play in Go \citep{silver2016mastering}, Starcraft 2 \citep{vinyals2019grandmaster}, DOTA 2 \citep{berner2019dota}, and poker \citep{brown2019superhuman} are seen as pivotal benchmarks in AI progress. While these feats were being accomplished, work was also being done showing the vulnerabilities of these methods. Past work also showed that policies are especially vulnerable against adversarial perturbations of image observations when white-box information is utilized \citep{huang2017adversarial}, and that an adversarial agent can easily learn a policy that can reliably win against a well-trained opponent in high-dimensional environments by learning from the rewards provided by the environment \citep{gleave2019adversarial}. These past findings, however, exploit information about the victim or environment that might not be realistic for the attacker's designer to know. 

Instead, we assume the attacker's designer knows as little information as possible, and ask the question: ``How successful can an attacker be when only observing the victim's actions without environment rewards?'' In particular, if the victim's reward function is $R_\nu$, then the attacker does not have access to $-R_\nu$. Would it be accurate to assume that no attacker could subvert a victim without such information? Here, we investigate one possible solution the attacker can utilize: a strategy for entropy maximization of both the attacker's and victim's policy. This solution proposes to first maximize the attacker's policy entropy to explore the victim's behavior, gathers the data from this exploration, and then applies batch RL to learn a new policy using the empirical entropy of the victim's policy as the reward function. This is intended to cause the victim's behavior to be more erratic. Since actions that maximize either the attacker's or victim's policy entropy at each state do not require direct knowledge of the environment's rewards, we call such actions \emph{reward-free attacks}.

\textbf{Contributions} We contribute (1) an explanation why strategically maximizing the attacker's and victim's policy entropy can negatively impact the victim, (2) an algorithm that is theoretically grounded in reward-free exploration \citep{jin2020reward}, and provide experiments that show (3) an attacker that maximizes victim policy entropy can negatively affect the victim. Hence, it is possible for an attacker to use relatively larger amounts of victim data to compensate for the designer's lack of prior environment or victim knowledge. To mitigate its potential negative impact, we propose some countermeasures on how to defend against such attacks.  These results underscore the necessity to reflect on what RL training standards are needed to ensure safe and reliable real-world RL systems even when victim motivation or environment information are unavailable to potential attackers.

In Section 2, we review past related work in adversarial RL and reward-free exploration. In Section 3, we justify the importance of reward-free attacks and propose our reward-free algorithm. In Section 4, we provide the results of experiments that show how reward-free attacks affect certain board games and multi-agent cooperative particle environments. In Section 5, we provide some conclusions, propose countermeasures, and suggest directions for future work.

\section{Related Work}

\label{gen_inst}

The work presented here attempts to further understand negative side effects in AI, which are one of the concrete problems in AI safety mentioned in \citet{amodei2016concrete}. Specifically, our work investigates reward-free attacks, which can deliberately increase the negative side effects a victim may encounter. This failure mode can also be seen as an example of adversarial optimization \citep{manheim2019multiparty}.

Some accomplishments have been made in introducing an adversarial element to the process of policy improvement in RL agents. Some examples include Robust Adversarial RL \citep{pinto2017robust}, and Risk Adverse Robust Adversarial RL \citep{pan2019risk}. There has also been research in RL that assumes an adversary that subverts a victim agent. As mentioned in the previous section, \citet{huang2017adversarial} use the victim's image observations to negatively affect its policy. \citet{gleave2019adversarial} showed that an adversary with access to environment rewards can quickly learn to defeat a trained victim. \citet{huang2019deceptive} and \citet{zhang2020Adaptive} use reward poisoning to trick the victim into learning a nefarious policy. Our work does not assume the attacker has the ability to manipulate the victim's observations, or the ability to poison the environment rewards the victim receives.

There is also the work by \citet{krakovna2020avoiding} on avoiding side effects in RL. Here, the setup and experiments are single-agent and minimizes side-effects by using auxiliary rewards that maximize potential rewards obtained from possible future tasks. Although our work is multi-agent and does not assume the antagonist has access to the environment's rewards, we hope to use their insights to extend future research in reward-free attacks.

We will model our multi-agent environment as a Markov game (\cite{shapley1953stochastic}, \cite{littman1994markov}) similar to what was defined in \citet{zhang2019non} where one agent has more information than the other. The foundation and algorithms are based on the reward-free RL framework by \citet{jin2020reward} and its extension using R{\'e}nyi entropy for exploration by \citet{zhang2021exploration}. The benefit of using this framework is that the theoretical guarantees hold for an arbitrary number of reward functions.

\section{Methods}
\label{headings}

The purpose of this paper is to (1) show that maximizing both the attacker's and the victim's policy entropy can be advantageous to an attacker, and (2) provide a reward-free RL algorithm that maximizes the victim's empirical policy entropy from the attacker's observations. In this section, we provide the intuition why maximizing victim policy entropy can be undesirable for a trained, static victim. Then, we introduce the reward-free RL algorithm and describe its theoretical benefits.

\subsection{Preliminaries}

We model the agents as a two-player, reward-free exploration Markov game. In this game, the victim ($\nu$) has access to the reward function $R_\nu$ while the attacker ($\alpha$) has no reward function. This is represented as $M = (\mathcal{S}, (A_\alpha, A_\nu), P, R_\nu)$ where $\mathcal{S}$ is the state set, $A_\alpha$ and $A_\nu$ are action sets, $R_\nu$ is the victim's reward function, and $P$ is the state-transition probability distribution. There are also the attacker's policy ($\pi_\alpha$) and the victim's policy ($\pi_\nu$). We hold the victim's policy fixed during the attacker's training. The motivation for this model is for the attacker to explore the victim's behavior while maximizing the entropy of its own policy $\pi_\alpha$ \citep{zhang2021exploration}. It then generates and collects trajectories while also keeping track of $\nu$'s action distribution at each state. Once the attacker $\alpha$ has enough data on the victim's behavior, it uses a batch RL algorithm (like Batch Constrained Q-Learning \citep{fujimoto2019off}), to learn a policy that maximizes $\nu$'s policy entropy.

Informally, a \textbf{reward-free attack} on a victim $\nu$, by attacker $\alpha$ (without knowledge of the rewards provided by the environment), can be described as a sequence of actions that (1) lower $\nu$'s expected returns, or (2) increase negative side-effects.

The following theorem can be seen as an explanation for what happens when an opponent lowers the victim agent's state-value over time. 

\begin{theorem}
Assume a finite, turn-based, zero-sum, deterministic game with no intermediate rewards and let $\nu$ have the static, optimal value function $V^{*}_{\nu}$. If $\nu$ wins, $0 < \gamma < 1$, $\nu$'s policy is greedy, and $n$ is the number of time steps (or game moves) left at state $s$ to traverse and win the game, then $n = \log_\gamma V^{*}_{\nu}(s)$. 
\end{theorem}

\begin{proof}
See supplementary material.
\end{proof}

This theorem states that in a board game, like Breakthrough or Connect-4, the number of steps left to win the game monotonically decreases as $V^{*}_\nu$ increases. This is useful because if the attacker is successful at lowering the value function of the victim, and the victim follows the optimal value function, the length of the game will increase. This is relevant to our investigation of reward-free attacks if we make the following assumption:

In board games like Breakthrough, Havannah and Connect-4, we measure negative side-effects by the average number of moves it takes to complete a game. This is motivated by the intuition that an expert player wants to win as quickly as possible. For example, in Breakthrough and Havannah, an attacker needs to know effective blocking strategies that prevent the victim from winning. The experiments in the next section will verify this theorem for some games. However, in games like Go, this assumption might not hold since an artificial agent in this game might not stop playing until all possible moves have been exhausted. For the appropriate games, like Breakthrough and Havannah, we use the average number of moves as evidence that the attacker is subverting the victim even if the attacker does not learn how to win the game.

\subsection{An Algorithm for Reward-Free Attacks}

\subsubsection{The Impact of Victim Entropy}

\begin{definition}
The \emph{victim $\nu$'s (policy) entropy} is $H^\nu(s)$. Hence, the attacker $\alpha$'s state-value function is 
\begin{equation}
    V_{\alpha}(s_t) = \mathbb{E}_{\pi_{\alpha}}[\Sigma^{\infty}_t \gamma^t H^{\nu}( s_t) | S_t = s_t]
\end{equation}
where $\gamma$ is the discount factor. Let $\pi_\nu(s)_i$ be the probability of action $i$ under policy $\pi_\nu(s)$. In our experiments, we either use Shannon entropy: $H^\nu(s) = -\sum^n_{i=0} \pi_\nu(s)_i \log \pi_\nu(s)_i$, or R{\'e}nyi entropy of order 0.5: $H^\nu(s) =  2 \log (\sum^n_{i=0} \pi_\nu(s)^{\frac{1}{2}}_i)$.
\end{definition}

For the rest of the paper, we will refer to victim policy entropy simply as \emph{victim entropy} when context is clear.

We believe, in most cases, a policy is useless if it is a uniform distribution at every state. Some real-world RL applications will require predictable behavior to be successful. There may exist situations where a uniform distribution over actions is not harmful, or even helpful to the agent. One example is rock-paper-scissors, where the Nash equilibrium is for all players to have a uniform distribution as their mixed strategy. It would be undesirable, however, for autonomous vehicles to have such policies. If an autonomous car were surrounded by people, you would not want the choices ``stop'' and ``go'' to have the same probability. There are environments that induce an inverse correlation between victim entropy and victim state-value. Taking board games as an example, there is the implicit assumption that, at certain states, the policy will only a have small number of actions to choose (low entropy) to achieve maximum cumulative rewards. If you want to quickly win in board games like Connect-4, the amount of moves that take you to the shortest winning path will decrease as you get closer to a winning state. Hence, an agent that maximizes victim entropy could learn how to avoid states where the victim will win with high certainty. 

While we do not imply that optimizing for victim entropy is better for the attacker to learn than environment rewards, we claim that it is possible to learn behavior that can subvert a victim by just observing their actions at each state. Hence, it is possible to use large amounts of victim observations to compensate for the attacker's lack of knowledge of the environment. In the experiment section, we show that maximizing victim entropy, combined with exploration maximizing R{\'e}nyi entropy on the attacker's policy, can successfully subvert victims that cannot be easily defeated by standard RL methods.

Now that we have explained why the impact of victim entropy can be harmful, we have motivation to construct an algorithm that maximizes the victim's empirical entropy derived only from observing the victim's actions.



\subsubsection{Combining R{\'e}nyi Entropy Exploration with Victim Entropy}

\begin{theorem}
In a Markov game $M$, with players $\alpha$, $\nu$ and corresponding discrete action sets $A_\alpha$, $A_\nu$. Let $\alpha$ have its policy $\pi_\alpha$ be the uniform distribution and $S_\alpha$ be the states $\alpha$ can reach following $\pi_\alpha$. Assume $\alpha$ can record $\nu$'s actions at each state. Then, for threshold $\epsilon > 0$, there exists an algorithm that approximately converges to the $\nu$'s victim entropy $H^\nu(s)$ for all $s \in S_\alpha$ within $\epsilon$.
\end{theorem}
\begin{proof}
See supplementary material.
\end{proof}

The implication that follows from this is the possibility that an agent could learn victim entropy from observing actions. That is, the empirical $\hat{H}^\nu(s)$ can be used as approximate rewards for an RL algorithm to train an agent to learn behavior that maximizes victim entropy. Hence, this algorithm will give a heuristic reward function for typical value-based methods. There are likely more efficient ways to accomplish this task, but the point is to show it is possible to create agents that can subvert victims through observation alone.

\begin{definition}
The \emph{victim's empirical (policy) entropy} $\hat{H}^\nu$ is the entropy of the empirical probability of the victim's action distribution at each state. Let $\hat{\pi}_\nu(s)_i$ be the empirical probability of $\nu$ taking action $i$ at state $s$. The victim's empirical Shannon entropy is: $\hat{H}^\nu(s) = -\sum^n_{i=0} \hat{\pi}_\nu(s)_i \log \hat{\pi}_\nu(s)_i$. The empirical R{\'e}nyi entropy of order 0.5 is: $\hat{H}^\nu(s) =  2 \log (\sum^n_{i=0} \hat{\pi}_\nu(s)^{\frac{1}{2}}_i)$.
\end{definition}

The following theorem is useful because it holds for any reward function:

\begin{theorem}[\citet{zhang2021exploration}]
 Let $d^\pi_h (s, a) := \Pr(s_h = s,a_h = a|s_1 \sim \mu;\pi)$, where $\mu$ is the initial state distribution and $H$ is the finite planning horizon and $h \in [H]$. Let $\omega$ be the set of policies $\{\pi^{(h)}\}^H_{h=1}$, where $\pi^{(h)}: S \times [H] \to  \Delta^{\mathcal{A}}$ and $\pi^{(h)} \in \argmax_\pi H_\alpha(d^\pi_h)$ (R{\'e}nyi entropy). Construct a dataset $\mathcal{M}$ with $M$ trajectories, each of which is collected by first uniformly randomly choosing a policy $\pi$ for $\omega$ and then executing the policy $\pi$. Assume

\begin{equation}
    M \geq c\left(\frac{H^2 S A}{\epsilon}\right)^{2(\beta + 1)} \frac{H}{A}\log \left(\frac{S A H}{p\epsilon}\right),
\end{equation}

where $\beta = \frac{\alpha}{2(1-\alpha)}$ and $c>0$ is an absolute constant. Then there exists a planning algorithm such that, for any reward function $r$, with probability at least $1 - p$, the output policy $\hat{\pi}$ of the planning algorithm based on $\mathcal{M}$ is $3\epsilon$-optimal, i.e., $J(\pi^{*}; r) - J(\hat{\pi}; r) \leq 3\epsilon$, where $J(\pi^*; r) = \max_\pi J(\pi; r)$.

\end{theorem}

 With some additional time for the attacker to observe the victim's actions at each state, we now have the following corollary:

\begin{corollary}

Under the conditions of Theorem 3, there exists a planning algorithm such that, if $\hat{H}^\nu$ is the reward function, then with probability at least $1 - p$, the output policy $\hat{\pi}$ of the planning algorithm based on $\mathcal{M}$ is $3\epsilon$-optimal, i.e., $J(\pi^{*}; \hat{H}^\nu) - J(\hat{\pi}; \hat{H}^\nu) \leq 3\epsilon$, where $J(\pi^*; \hat{H}^\nu) = \max_\pi J(\pi; \hat{H}^\nu)$.
\end{corollary}

Using the reward-free RL framework, there are now some guarantees for any potential designers intending to train agents that learn reward-free attacks.

The outline of our algorithm is as follows:
\begin{itemize}
    \item \textbf{Exploration}: With a replay buffer of size $n$, while playing against victim $\nu$, learn a value function $V_\theta$. At each state $s_t$, the reward for the attacker is the R{\'e}nyi entropy:  $r(s_t) = H_{\alpha}(\pi_\theta(s_t))$ where $\pi_\theta$ is the softmax of the q-value function $Q_\theta$. 
    \item \textbf{Collect data}: Rollout policy $\pi_\theta$ against the victim $\nu$. Record $M$ number of trajectories and $K$ number of victim actions for the victim's empirical entropy.
    \item \textbf{Planning}: With the victim's empirical entropy as the reward function, use a batch RL algorithm to learn the attacker's policy.
\end{itemize}

We show in the next section that the combination of the attacker maximizing its own policy entropy for exploration, and then using the victim's empirical entropy as a heuristic reward function, can be successful in subverting trained victims.

\section{Results}

\subsection{Experimental Setup}

Here, we explain our experimental setup\footnote{Further experimental setup details are provided in the supplementary material.} and how it provides answers to the questions mentioned in the introduction. To show the effect of victim entropy, we use value-based RL for board games and policy-based RL for cooperative navigation. Then, we show our results for the reward-free algorithm, which show that the combination of maximizing attacker and empirical victim entropy can learn reward-free attacks.

\subsubsection{Other Agent Types}

To further illustrate how an agent that learns from victim entropy apart for other possible agents, we introduce other types of agents to compare and contrast with the reward-free attacker. One agent is the \textbf{antagonistic value} agent that has the reward function $R(s_t) = -V_\nu(s_{t+1})$, where $V_\nu(s)$ is $\nu$'s state-value function. We consider this agent to be a type of ``cheater'' that has direct access to the victim's value function so that the agent can minimize it. Specifically for board games, the other agent is the \textbf{move maximizing} agent that is $R(s_t) = m_t$, where $m_t$ is the number of moves it makes in the game at time $t$. This agent exploits the fact that greedy agents take the shortest route to win the game. Hence, it is a heuristic that requires knowledge about the environment. The reason we introduce these other types of agents is to compare the performance of the attacker that maximizes policy entropy to agents that require more knowledge about the victim or environment.

\subsubsection{Board Games}
\begin{figure}[t]
  \centering
  \includegraphics[scale=0.23]{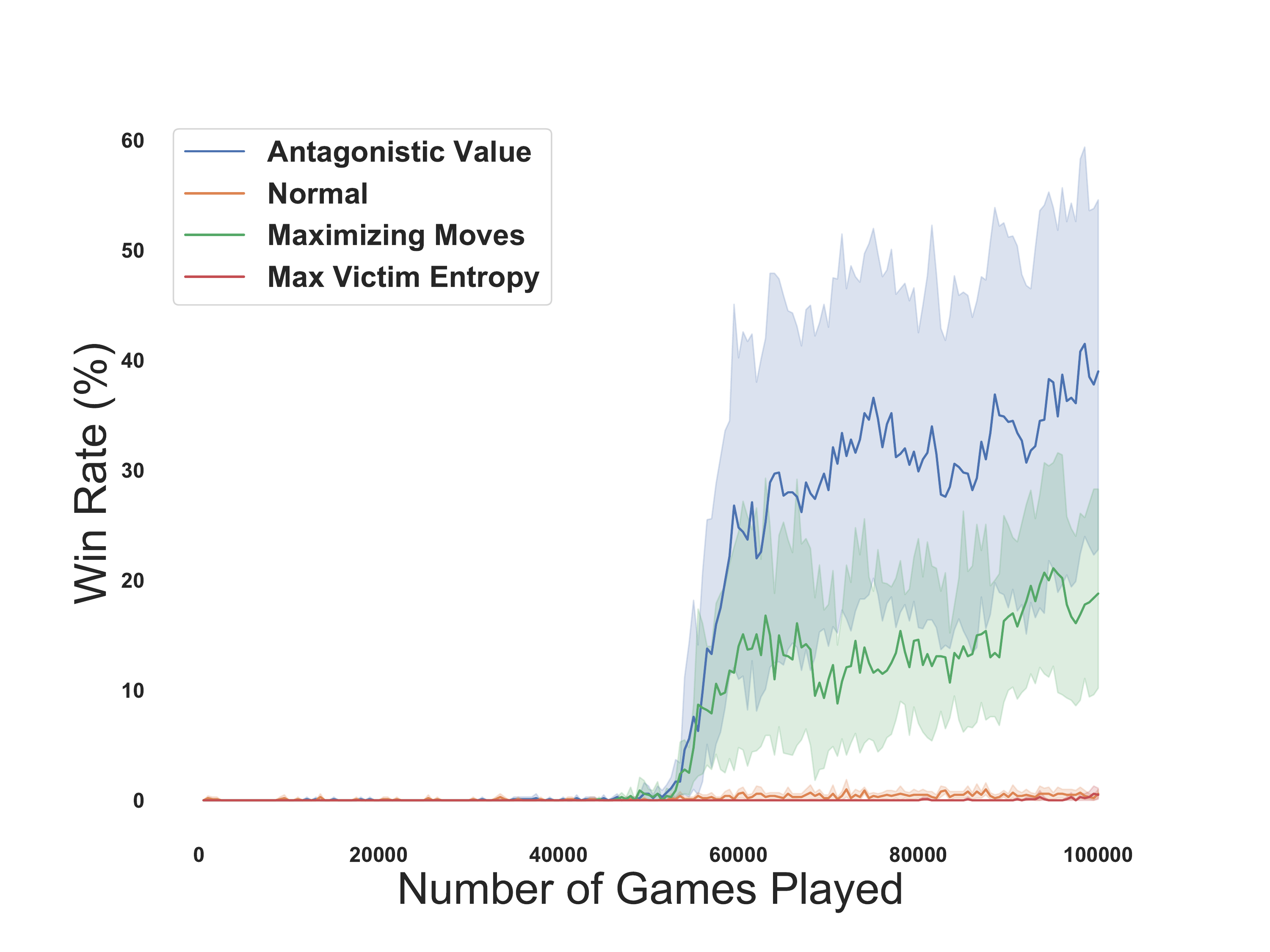}
  \includegraphics[scale=0.23]{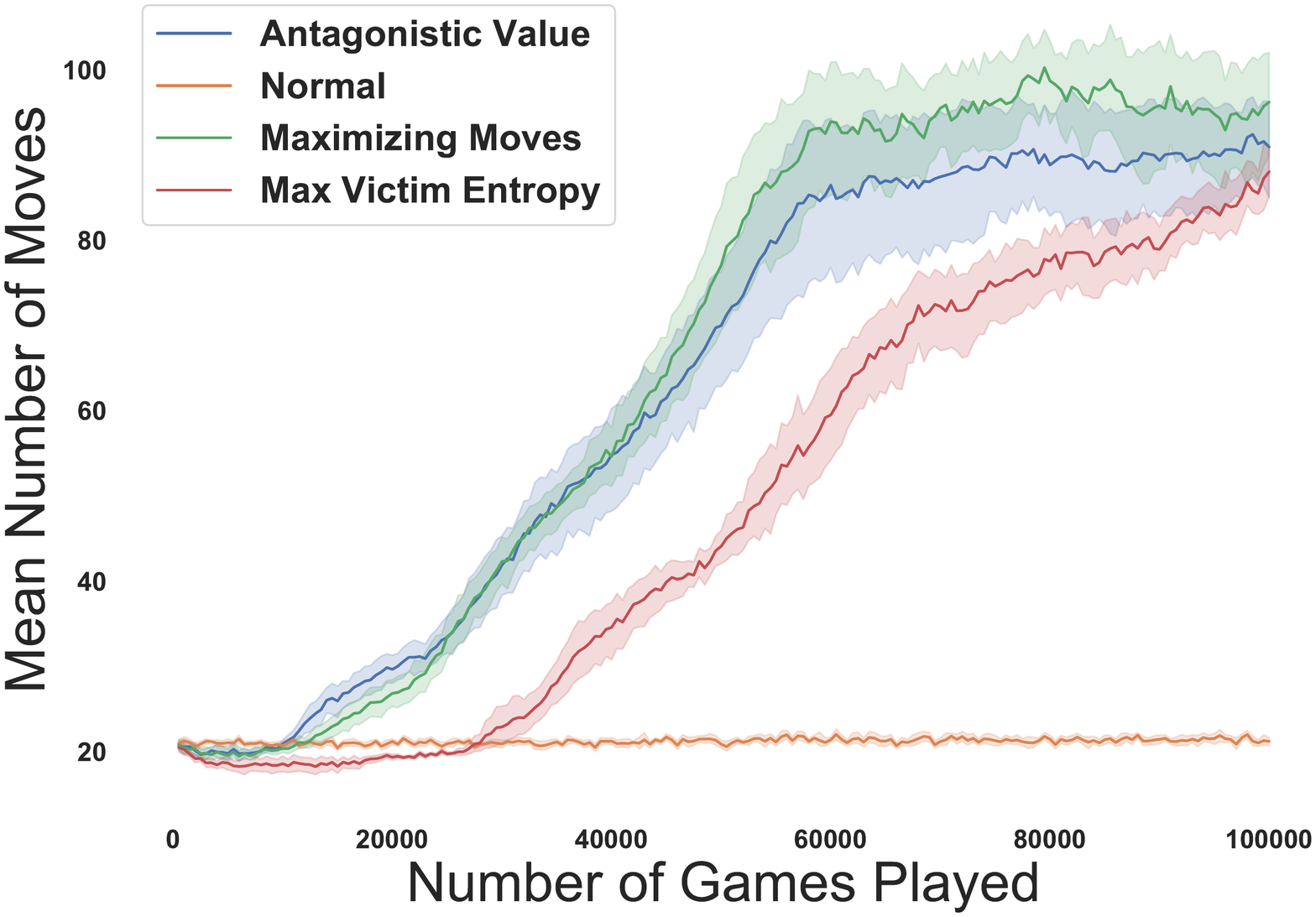}
  \caption{Breakthrough Deep Q-Learning Training Results: \\
  \textbf{Top}: While the normal agent and victim entropy agent have win rates consistently close to 0\%, the antagonistic value agent ends with a mean win rate of around $40 \pm 15\%$. The antagonist that learns from maximizing moves is slightly below $20 \pm 10\%$.\\
  \textbf{Bottom}: The normal agent mean number of moves is consistently around 21. The antagonistic value agent ends with a mean number of moves of around $90 \pm 5$. The agent that maximizes moves is slightly higher than the antagonistic value agent. The most surprising result was the victim entropy agent learning to increase the number of moves while never learning to win against the victim.}
  \label{fig:breakthrough}
\end{figure}

\begin{figure}[t]
  \centering
  \includegraphics[scale=0.23]{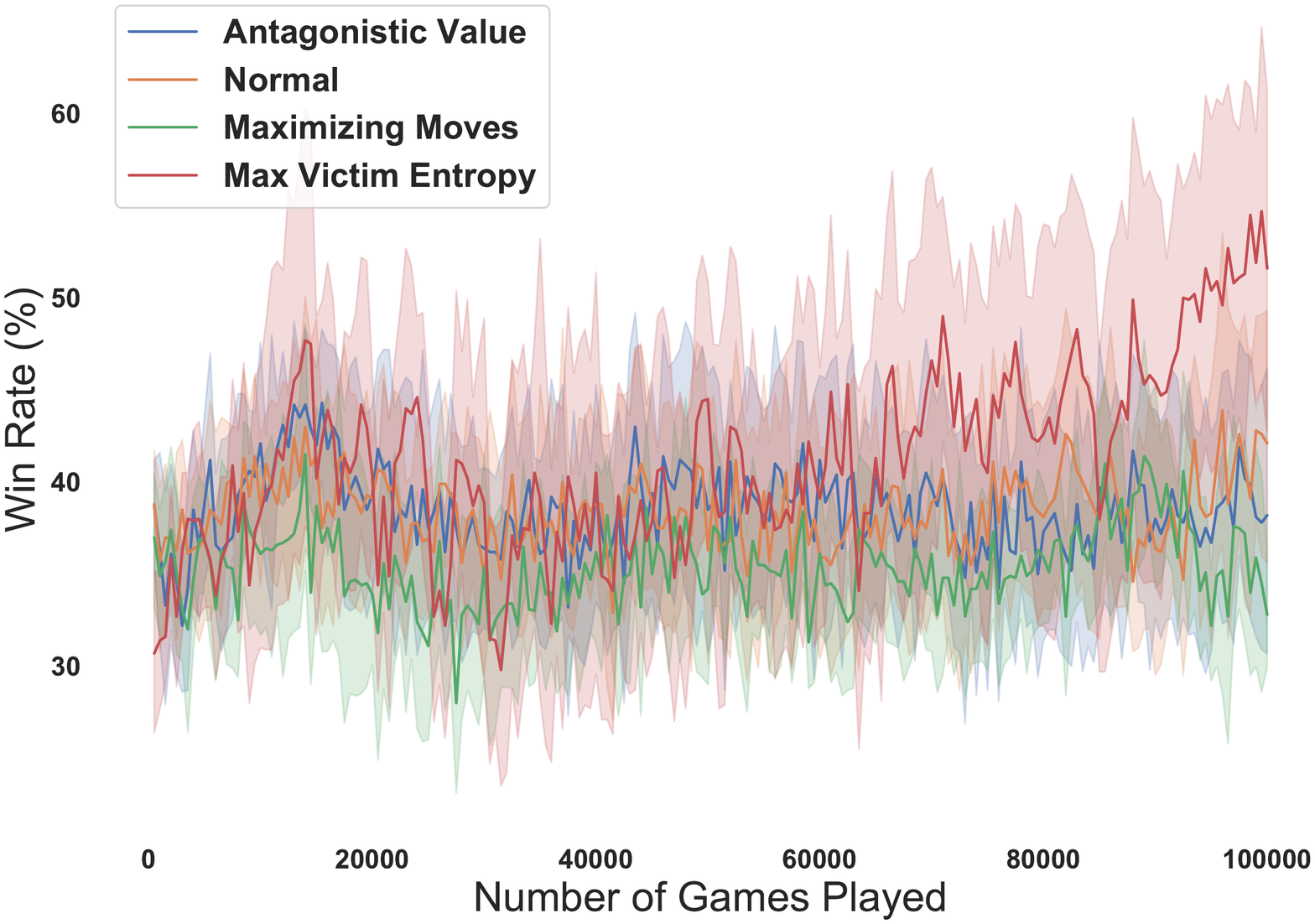}
  \includegraphics[scale=0.23]{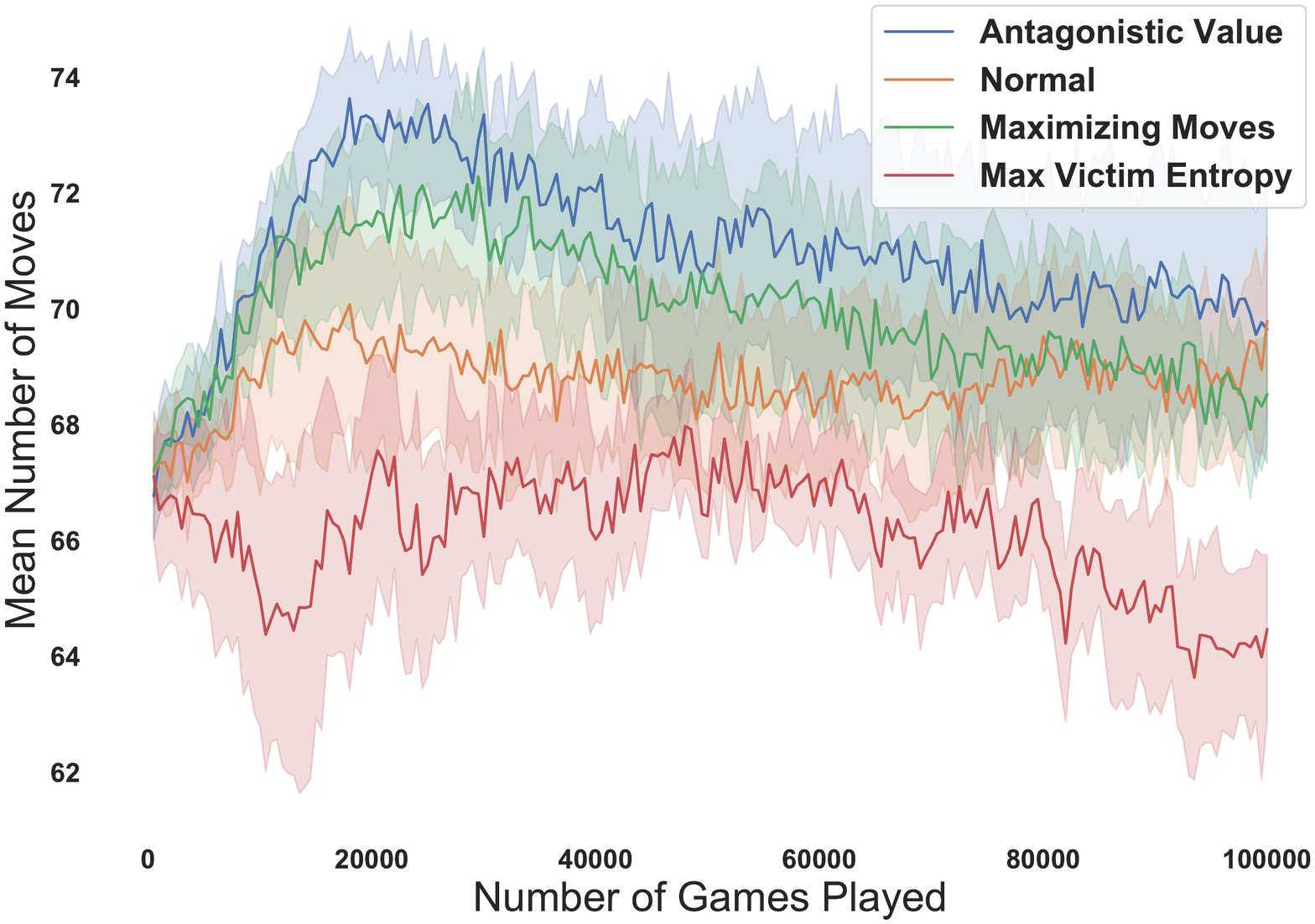}
  \caption{Havannah Deep Q-Learning Training Results: \\
  \textbf{Top}: The victim entropy agent ends with a mean win rate of around $50 \pm 10\%$ While the other agents have a win rate consistently close to 40\%. \\
  \textbf{Bottom}: The victim entropy agent ends with a mean number of moves of around $64 \pm 2$. The other agents' mean number of moves is consistently around 70.}
  \label{fig:havannah}
\end{figure}

We train the attacker and victim using the game environment OpenSpiel\footnote{\url{https://github.com/deepmind/open_spiel} (Apache-2.0 License)} \citep{lanctot2019openspiel}. Both agents' policies are deep Q-networks that are trained in a manner similar to \citet{mnih2015human}. Given the non-transitive, cyclic nature of player strength in real-world games \citep{czarnecki2020real}, it is difficult to give an objective ranking to our victim agents. Given the results in Figures \ref{fig:breakthrough} and \ref{fig:havannah}, we assume that the Breakthrough victim is the stronger agent ($\sim100\%$ win rate against Deep Q-Learning agent) compared to the Havannah victim ($\sim60\%$ win rate against Deep Q-Learning agent). This is likely because Breakthrough is a relatively easier game for a Deep Q-Learning agent to find an optimal policy.

\subsubsection{Multi-agent Particle Environment}

We also investigate the effect of maximizing victim entropy in environments with more than two agents. In particular, we use the cooperative navigation scenario in the OpenAI Multi-Agent Particle Environment\footnote{\url{https://github.com/openai/multiagent-particle-envs} (MIT License)} \citep{lowe2017multi}. Here, we train 3 agents at a time using multi-agent deep deterministic policy gradient (MADDPG) \citep{lowe2017multi}. We train 3 separate groups of agents: (1) a group that learns from the rewards of the environment, and a group with one attacker agent that focuses on one victim agent by learning (2) value-based antagonistic behavior or (3) maximizing victim entropy. In our experiments, we let the trained group of normal MADDPG agents interact for 500 time steps. After this, we replace one of the normal agents with one of the antagonistic agents and observe the group rewards. We also plot the behavior of the normal group without inserting an attacker agent as a control group. The point of this experiment is to see if introducing an attacker can subvert cooperative group behavior. This provides some idea of what might happen if the architect inserts an attacker $\alpha$ into groups of agents similar to the victim $\nu$ it was training against. 

\subsection{Victim Entropy Results}

In this subsection, we assume the attacker has access to the victim's policy. These results reflect what happens when the attacker is able to observe enough of the victim's actions so that the empirical entropy is close to the true victim policy entropy.

In some board games, learning from rewards or victim entropy can be more effective strategies in achieving higher win rates. This is surprising if you consider access to the victim's state-value function as a form of cheating. Against stronger agents, however, antagonistic value seems to perform better compared to the other antagonists. In Figure \ref{fig:breakthrough}, it performs the best against the strong Breakthrough victim. In Figure \ref{fig:havannah}, it produces as many moves as the normal agent but does not seem to win any more than the other agents. 

In Figure \ref{fig:havannah}, the agent that maximizes victim entropy achieves a win rate similar to the DQN agent that learns from rewards. In Figure \ref{fig:breakthrough}, where it learns to increase the amount of moves while maintaining a near-zero win rate. An explanation regarding how these results seem to contradict each other could be that the Breakthrough victim is stronger because it is a simpler game to learn for a Deep Q-Learning agent. Since all victim agents train with the same amount of games for all board games, it is probably easier for an attacker to learn how to subvert a Havannah victim since it is a harder game for a Deep Q-Learning agent to master. In some ways, the attacker's performance in Breakthrough can be interpreted as the closest to purely antagonistic behavior because it is clearly not learning how to win but seems to learn how to increase negative side-effects, which makes it a reward-free attack.

\begin{figure}[ht!]
  \centering
  \includegraphics[scale=0.2]{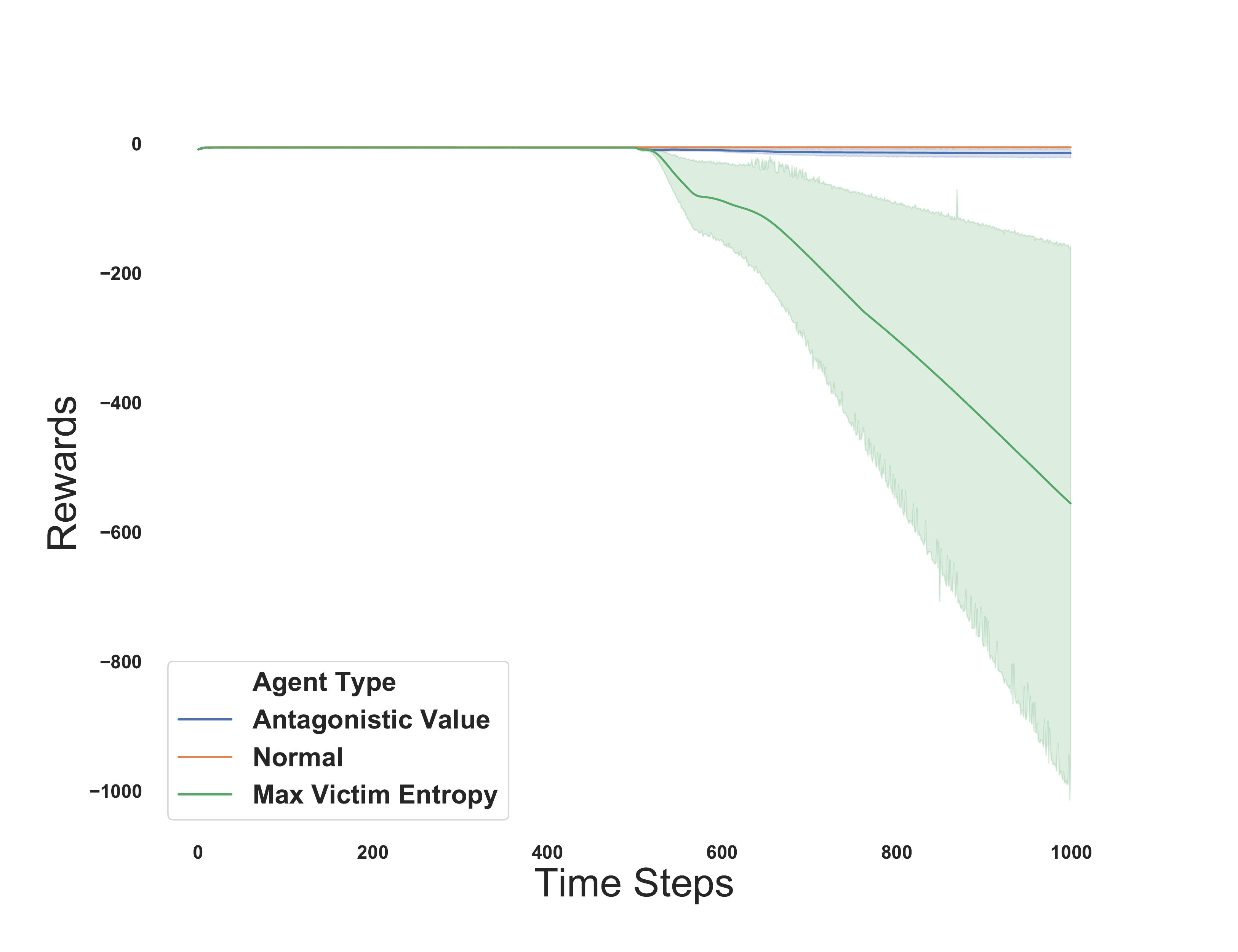}
  \includegraphics[scale=0.2]{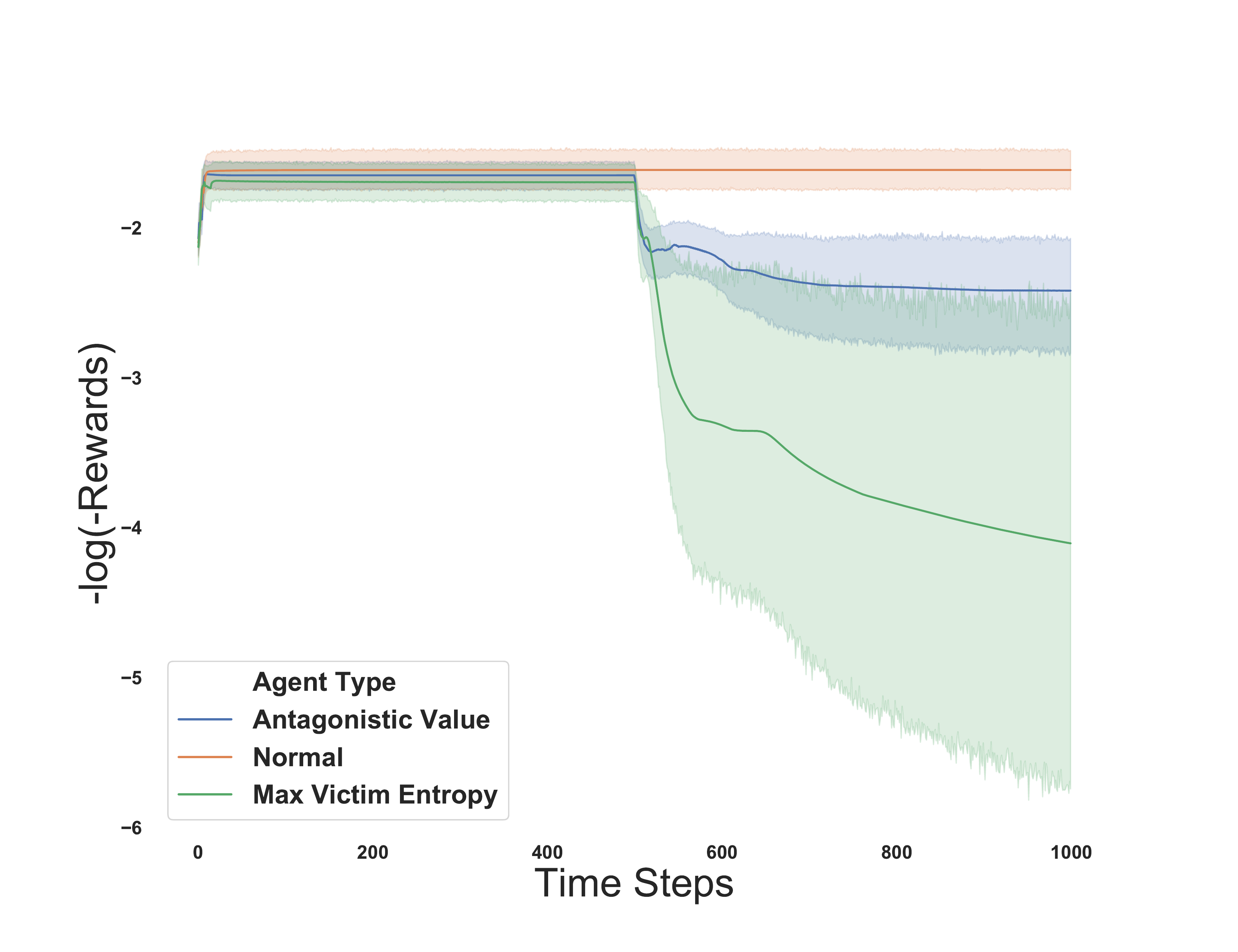}
  \caption{Cooperative Navigation Results: \\
  An antagonistic agent is inserted at time step 500 in the cooperative navigation scenario of the particle environment. The orange line shows the group behavior if no antagonistic agent is inserted.\\
  \textbf{Top}: When the max victim entropy agent is inserted, the decrease in returns is more significant than both the antagonistic value and normal groups.\\
  \textbf{Bottom}: This is the same as the plot above except the y-axis has been transformed by the function $-\log(-x)$. This makes it easier to compare the performance of all three groups of agents together.}
  \label{fig:maddpg}
\end{figure}

In cooperative navigation, learning to minimize one of the victim's state-value has a noticeable effect. In Figure \ref{fig:maddpg}, inserting the antagonistic value agent produces a decrease in the group's rewards. However, the mean of this decrease stabilizes around $-14.0 \pm 5.0$. In all experiments performed, this attacker is successful, without necessarily learning to win, in subverting the victim in some way. 

Another surprising result is when an victim entropy agent is inserted into a cooperative navigation group. In Figure \ref{fig:maddpg}, not only is the mean of the rewards constantly decreasing, the variance seems to be increasing too. This is in stark contrast to the antagonistic value agent, where the trend in both the mean and variance of the rewards plateaus as the time steps increase. Although the antagonistic value agent has a negative effect on the group, the plots in Figure \ref{fig:maddpg} show that this effect seems less significant when comparing it to the victim entropy agent's impact.

In these experiments, the attacker has access to the victim's policy, which was the softmax of the action-values at the current state, and calculates the entropy from this policy. Outside of Assumption 3, it is unclear why victim entropy is notably effective in these situations. The success of this method underscores the importance of just knowing how your opponent will act at a given state.

\subsection{Reward-Free Algorithm Results}

\begin{figure}[t]
  \centering
  \includegraphics[scale=0.23]{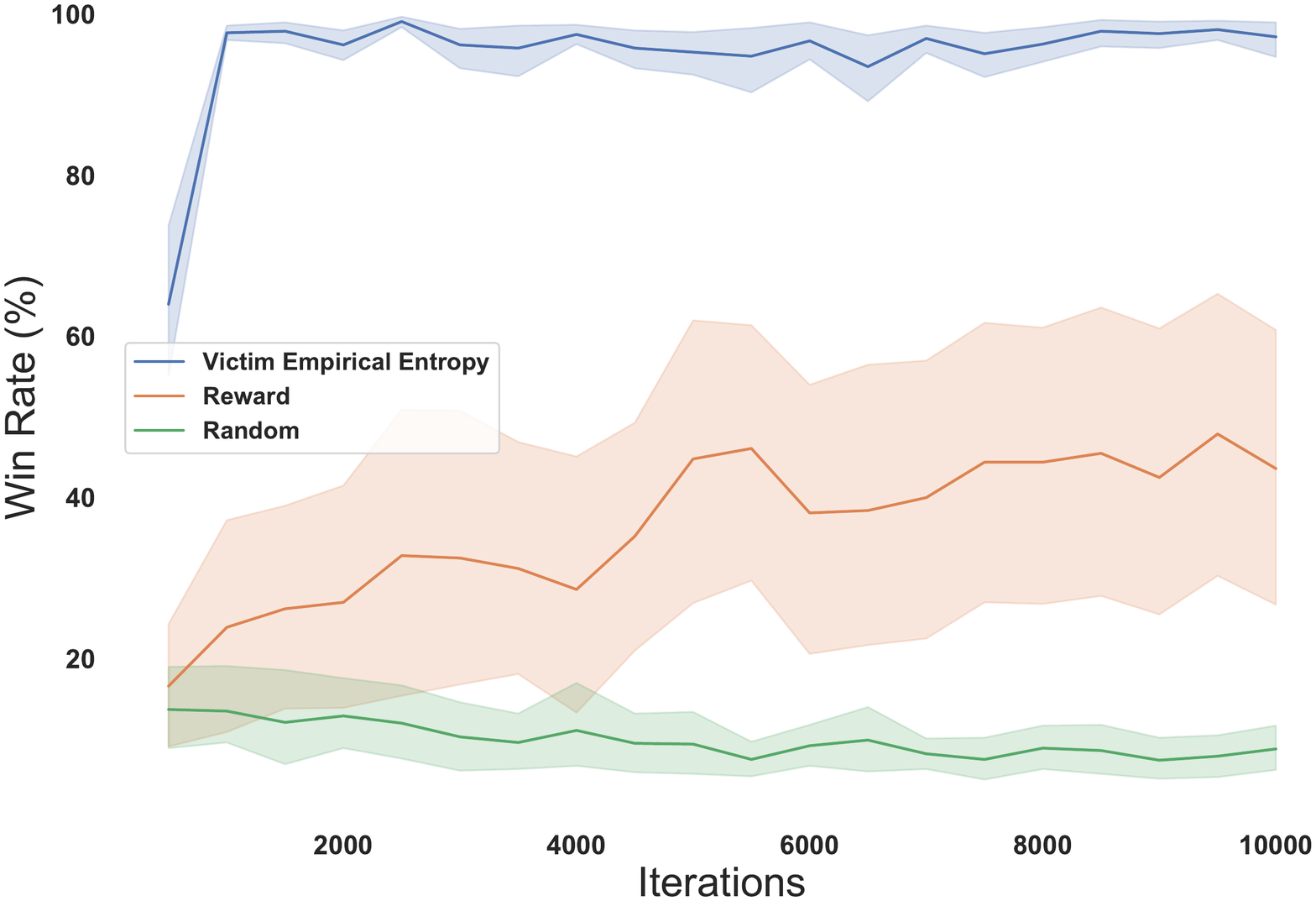}
  \includegraphics[scale=0.23]{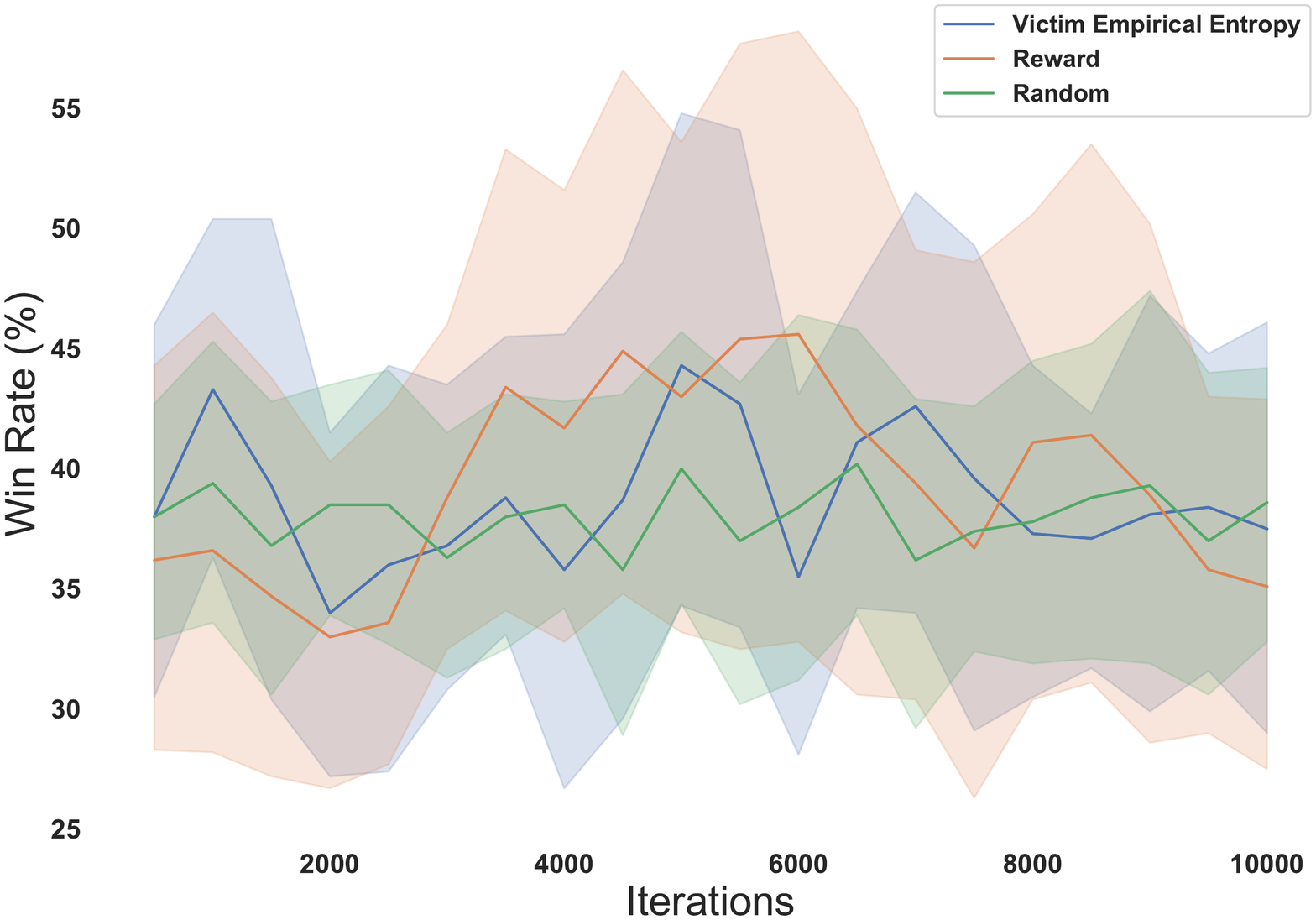}
  \caption{Batch RL Training Results: \\
  \textbf{Top (Breakthrough)}: While empirical victim entropy agent (blue) is consistently close to 100\%, the agent that learns from game rewards (orange) ends with a mean win rate of around $40 \pm 20\%$. The agent that learns from a random reward function (green) performs around $10\%$.\\
  \textbf{Bottom (Havannah)}: All agents (empirical victim entropy, game reward, random) perform about the same.}
  \label{fig:batchRL}
\end{figure}

In these experiments, we follow the reward-free exploration algorithm described in the methods section with R{\'e}nyi entropy of order 0.5. Here, the attacker no longer has access to the victim's policy and only observes the victim's actions at each state. In the rollout stage, the policy generated $10^5$ $(state, action, next$ $state, reward)$ tuples and $10^6$ victim actions for the empirical victim entropy. Lastly, the agent uses a batch (offline) version of the deep Q-learning algorithm used in the previous victim entropy results. We also trained a groups of agents on the game reward function, and a random reward function defined as a uniform distribution over game rewards $(-1, 0, 1)$ at each state.

The results in Figure \ref{fig:batchRL} seem counterintuitive. The attacker was able to immediately learn near-optimal behavior against the stronger victim in the game of Breakthrough that surpasses even the antagonistic value agent (Figure \ref{fig:breakthrough}). Although we do not have a clear, intuitive explanation for this result, we speculate that victim entropy provides a useful signal when there are sparse rewards. During the exploration phase, the attacker maximizes expected discounted R{\'e}nyi entropy to obtain a diverse dataset. We observed, however, that during the rollout phase, the policy generated a very small amount of winning trajectories (less than 1\% win rate). This might explain the why using game wins/losses as the reward function performed worse than using empirical victim entropy. It is still unclear why this combination of attacker/victim policy entropy was so effective. As we discussed previously, one possible explanation is that this victim's value function has a strong inverse correlation with the entropy of its policy. This result, where an attacker finds a weakness in the victim, is analogous to classifiers that achieve high accuracy on in-distribution examples but are not robust to certain adversarial attacks \citep{hendrycks2020many}.

The attacker was not able to learn how to reliably win against the weaker Havannah agent.  We believe that, due to the relative difficulty of the game, this weaker victim has never really learned how to win but to "guess" which trajectories will win about 60\% of the time against a standard Deep Q-Learning agent. Hence, the information from observing this victim's actions is less useful than from observing the actions of stronger agents.

\subsection{Limitations}
Some limitations of these experiments include (1) the environments used are much simpler than more ``real-world'' environments (e.g. driving simulations with pedestrians), (2) the assumption of perfect information, and (3) and the lack of theory on the impact of entropy in RL. Future research could include more complex environments that are closer to real-world RL applications that assumes either the victim or the attacker might not know the underlying state (e.g. POMDPs \citep{kaelbling1998planning}). While there is work that tries to understand the empirical impact of entropy in RL \citep{ahmed2019understanding}, we are unaware of a theoretical framework that adequately explains entropy's beneficial/detrimental effect on the intelligent behavior of multiple agents. Hence, we instead provided some intuition and experiments that explain the impact of using victim entropy as a reward function.

\section{Discussion}

Our work is inspired by past research in adversarial machine learning in that it exposes a problem that may arise when a malicious actor learns strategically from no prior knowledge of the victim's motivation. The problem we attempt to model is when a human designer builds an attacker to compete against a trained victim agent that learns only through observing the victim's actions. Specifically, we showed that an attacker maximizing the victim’s policy entropy can be an effective heuristic. The most surprising result of our work was the effectiveness of both maximizing R{\'e}nyi and victim entropy in stages, and the implications of this strategy. Our results demonstrate the effects of optimizing for the victim's disorder. The framework of reward-free exploration in RL shows that it is possible to learn such behavior without access to the environment rewards, value function, or policy. All you would need to do is observe the victim's actions at each state. These results, along with the impact of entropy in policy optimization in \citet{ahmed2019understanding}, show the importance of entropy in RL and the need to further our understanding of it.

We propose a number of possible defenses from these attacks that would benefit from further research:\\

1. Use agents that learn reward-free attacks as a test adversary against agents intended to be deployed for important tasks. An agent that is exploited by such an attacker could be subverted in real-world settings.\\
2. Never keep an agent static. The reward-free attacks presented here assume the victim has a stationary policy. A victim that continues to learn might make the attacker's observations less useful.\\
3. Have a diversity of policies. If possible, have different RL algorithms learn their own policies and switch when appropriate. The attackers presented here assume there is only one victim policy instead of many. 

Another direction, related to AI safety and cybersecurity, would be to use the notion of reward-free attacks as a more realistic model for adversarial RL than agents with access to the same environment rewards as its victim. Other directions include investigating the dynamics of such agents or developing methods that can better detect such behavior. Lastly, our definition of reward-free attacks encapsulates how a malicious designer, without knowledge of how the victim learned from the environment, would intend to create an attacker. This follows from the intuition that agents require some level of certainty when choosing actions at each state, which motivates potential attackers to choose actions that cause their victims to be less certain. In situations where this intuition reveals itself to be inaccurate, further research into why this intuition fails is worth investigating.
\bibliographystyle{plainnat}
\bibliography{ref}

\appendix
\onecolumn

\section{Theorems and Proofs}

\begin{theorem}
Assume a finite, turn-based, zero-sum, deterministic game with no intermediate rewards and let $\nu$ have the static, optimal value function $V^{*}_{\nu}$. If $\nu$ wins, $0 < \gamma < 1$, $\nu$'s policy is greedy, and $n$ is the number of time steps (or game moves) left at state $s$ to traverse and win the game, then $n = \log_\gamma V^{*}_{\nu}(s)$. 
\end{theorem}
\begin{proof}
Assume at the end of the game, the winning agent gets 1 point and the losing agent gets -1 point. From these assumptions, $V^{*}_{\nu}(s_{win}) = 1$, where $s_{win}$ is the winning state for $\nu$. \\
If $i$ is the number of steps left to traverse at state $s$, it suffices to show that $V^{*}_{\nu}(s) = \gamma^i$.
This can be proven by induction on the number of steps left for the greedy agent to reach the winning terminal state using the optimal value function.\\
Base case: If there is 1 step left, the greedy agent is at state $s$ such that $V^{*}_{\nu}(s) = 0 + \gamma * V^{*}_{\nu}(s_{win}) = \gamma * 1 = \gamma^1$.\\
Inductive case: Assume  $V^{*}_{\nu}(s') = \gamma^i$, for all states $s'$ such that $i$ is the number of steps left. Let $s$ be a state such that the greedy policy chooses the action that leads to $s'$ from $s$ and there are $i+1$ steps left to traverse. Since there are no intermediate rewards, we have $V^{*}_{\nu}(s) = 0 + \gamma V^{*}_{\nu}(s') = \gamma * \gamma^i = \gamma^{i+1}$. \\
Hence, if $i$ is the number of steps left to traverse at state $s$, $V^{*}_{\nu}(s) = \gamma^i$, which implies $i = \log_{\gamma}V^{*}_{\nu}(s)$.
\end{proof}

\begin{theorem}
In a Markov game $M$, with players $\alpha$, $\nu$ and corresponding discrete action sets $A_\delta$, $A_\nu$. Let $\alpha$ have its policy $\pi$ be the uniform distribution and $S_\alpha$ be the states $\alpha$ can reach following $\pi$. Assume $\alpha$ can record $\nu$'s actions at each state. Then, for threshold $\epsilon > 0$, there exists an algorithm that approximately converges to the $\nu$'s victim entropy $H^{\nu}(s)$ for all $s \in S_\delta$ within $\epsilon$.
\end{theorem}
\begin{proof}
Let $S_\delta \subseteq \mathcal{S}$ be the set of all states the antagonist can reach through following a random policy in a particular environment. Let $\mathcal{T}$ be a hash table with $s \in S_\alpha$ as the keys, and the values as vectors of zeros [0, 0, \dots, 0] that are the length of available actions to the victim at the corresponding state. We also need hash tables $\mathcal{H}_0$ and $\mathcal{H}_1$ both with $s \in S_\alpha$ as the keys. For $\mathcal{H}_0$, the table of previous victim entropies, each key $s \in S_\alpha$ has a corresponding value 0. For $\mathcal{H}_1$, the table of current victim entropies, each key $s \in S_\alpha$ has a corresponding value $\epsilon$. Let $d(\mathcal{H}_0, \mathcal{H}_1)$ be the difference between $\mathcal{H}_0$ and $\mathcal{H}_1$ defined as:

\begin{equation*}
d(\mathcal{H}_0, \mathcal{H}_1) =  \sum_{s \in S_\alpha} \Big| \mathcal{H}_0[s] - \mathcal{H}_1[s] \Big|
\end{equation*}

\begin{algorithm}
Initialize hash table $\mathcal{T}$ such that for each key $s \in S_\alpha$, there is a corresponding vector of zeros [0, 0, \dots, 0] that are the length of the number of available actions as the value\\
Initialize hash table $\mathcal{H}_0$ such that for each key $s \in S_\alpha$, the corresponding value is 0\\
Initialize hash table $\mathcal{H}_1$ such that for each key $s \in S_\alpha$, the corresponding value is $\epsilon$\\
Assume $\pi^{(t)}(s) = \argmax_\pi d^\pi_t(s)$\\
\While{not $d(\mathcal{H}_0, \mathcal{H}_1) < \epsilon$}{
Initialise state $s_1$\\
    \For{t = 1, T}{
        \If{h > 1}{Update $\mathcal{H}_0[s_{t-1}] := \mathcal{H}_1[s_{t-1}]$}
    Take action according to policy $\pi^{(t)}$\\
    Observe $\nu$'s action $a_t$\\
    Add +1 to the corresponding action index $i$ in the vector at $\mathcal{T}[s_t]_i$\\
    Calculate probabilities of action distribution $p^t_i = \dfrac{\mathcal{T}[s_t]_i}{\sum_j T[s_t]_j}$ for each action $i$\\
    Update $\mathcal{H}_1[s_t] := -\sum_i p^t_i\log(p^t_i)$\\
    }
    
}
\Return $\mathcal{H}_1$
\label{alg:ent}
\caption{Learning Victim Entropy in $S_\alpha$}
\end{algorithm}

Hence, if at least one state $s \in S_\alpha$ has not been explored, we have $d(\mathcal{H}_0, \mathcal{H}_1) \ge \epsilon$. Algorithm 1 (next page) can then be used to output $\mathcal{H}_1$ as a table of victim entropies for all $s \in S_\alpha$.

\end{proof}


\section{Experiment Methodology}

\subsubsection{Board Games}
We train the antagonist and victim using the game environment OpenSpiel. Specifically, the agents use $\epsilon$-greedy policies with 6-layer or 7-layer linear neural network value functions implemented in PyTorch. The victims are trained first over 500,000 games of Breakthrough, Havannah, and Connect-4 against random agents. These games were chosen because it was easy for RL agents to receive reward signals against random agents. We also tried using methods of self-play, but none of these agents could consistently win. Our best agent turned out to be the Breakthrough agent that was trained against a random agent.  Against random agents, each victim agent reaches around $95\%$ to $99\%$ win rate. We then train and measure the performance of 10 agents learning of each type (victim entropy maximizer, antagonistic value, and move maximizer) from the same victim agent. 

Given the non-transitive, cyclic nature of player strength in real-world games, it is difficult to give an objective ranking to our victim agents. Hence, we assume that the Breakthrough victim is the strongest since it achieved highest win rate ($\sim100\%$ win rate) when other Deep Q-Learning agents trained against it, compared to the weaker Havannah victim ($\sim60\%$ win rate).

\subsubsection{Multi-agent Particle Environment}

We also investigate the effect of antagonistic behavior in environments with more than two agents. In particular, we use the cooperative navigation scenario in the OpenAI Multi-Agent Particle Environment. In this scenario, the reward function for each agent is the negative sum of (1) the minimum distance over all the euclidean distances between the agents and their corresponding nearest landmark, plus (2) the number of times the agent collided with another agent. Here, we train 3 agents at a time using multi-agent deep deterministic policy gradient (MADDPG). For 2,000,000 time steps, we train 3 separate groups of agents: (1) a group that learns from the rewards of the environment, and a group with one attacker agent that focuses on one victim agent by learning (2) value-based antagonistic behavior or (3) maximizing victim entropy.

In our experiments, we let the trained group of normal MADDPG agents interact for 500 time steps. After this, we replace one of the normal agents with one of the antagonistic agents and observe the group rewards. We also plot the behavior of the normal group without inserting an attacker agent as a control group. The point of this experiment is to see if introducing an attacker can subvert cooperative group behavior. This provides some idea of what might happen if the architect inserts an attacker $\alpha$ into groups of agents similar to the victim $\nu$ it was training against. For each group (normal, antagonistic value, victim entropy), we run the experiment 10 times each to better measure group behavior.

\subsection{Evaluation Methodology}

To evaluate the attacker trained using Deep Q-Learning, we catalog the performance a group of 10 agents with their own random seed as each agent in the group separately trains over many games. However, when evaluating two trained game-playing agents with DQN policies, it is not immediately obvious how to measure aggregate performance during training. During intermediate evaluation while training, $\epsilon$-greedy agents need to set $\epsilon$ to 0 to avoid arbitrary randomness that does not contribute to the measurement of performance. On the other hand, if we set $\epsilon$ to 0 in games where the first state is the same in every game, a greedy policy is just a function from states to actions. That is, the greedy policy makes the same move for every state. Hence, two greedy agents playing Go or Breakthrough from the very beginning will just make the same moves every game. This makes it impossible to evaluate the agent's skill over a diverse set of game scenarios.

We propose the following agent evaluation methodology. Every 500 games during the training of the antagonistic agent, the method we devised is the following:
\begin{enumerate}
    \item Start a new game between the victim agent and a random agent.
    \item Have the agents play a constant number of moves (typically 5-10 moves for each agent, depending on the game).
    \item When the constant number of moves is reached, replace the random agent with the antagonist. Have both the victim and antagonist agents continue playing until the end of the game.
    \item Record the winner and number of moves.
    \item Repeat process until you reach the amount needed for your sample size (typically 100 games).
\end{enumerate}

This tests the ability of the agent's greedy policy to pick the actions that maximize antagonistic behavior over many different scenarios without forcing $\epsilon$ to be nonzero. Hence, the agents will always pick what they consider to be the best action at the current state.


\subsection{Additional Notes on Maximizing Entropy}
During the exploration phase, the entropy of the terminal states is zero. This fits the intuition that, at the end of the game, the only valid move for the players is to stop playing. As the training during the exploration phase progressed, we observed that the mean number of moves increased over time. This provides evidence that the agent is adequately exploring the state-action space. During the planning phase, all states that were not generated during the rollout phase were given a reward of -1. This is because the empirical victim entropy at each state is valid only if at least one action by the victim was observed. We chose -1 so that the policy derived from the batch RL algorithm would learn to avoid states that were not explored in the rollout phase. The MADDPG attacker learns from Shannon entropy. All other agents learn from R{\'e}nyi entropy of order 0.5.

\section{Implementation and Hyperparameter Details}

Experiments were carried out on Dual Intel Broadwell E5-2620 v4 @ 2.10GHz CPU, with either a Dual NVIDIA P100 12GB PCI-e based GPU or Dual NVIDIA V100 16GB PCI-e based GPU.\\

\begin{table}[h]
    \centering
    \begin{tabular}{ |p{3.5cm}||p{3.5cm}|}
     \hline
 \multicolumn{2}{|c|}{} \\
 \hline
 Number of Hidden Layers & 6 (for Go and Connect-4)\\
 & 7 (for Breakthrough and Havannah)\\
 & 4 (for MADDPG agents)\\
 Number of Units   & 256 (for DQN agents)   \\
 & 64 (for MADDPG agents)\\
 Activation Function Used After Each Hidden Layer&   ReLU (for DQN agents)\\
 Batch Normalization Applied After Activation Function &True (for DQN agents) \\

 \hline
    \end{tabular}
    \caption{Neural Network Architecture Hyperparameters}
    \label{tab:nnhyper}
\end{table}


\begin{table}[h]
    \centering
\begin{tabular}{ |p{3.5cm}||p{3.5cm}|}
 \hline
 \multicolumn{2}{|c|}{} \\
 \hline
 Optimizer & Adam\\
 Learning Rate & $10^{-4}$\\
 Discount Factor $\gamma$   & 0.9 (for game or random rewards)\\
 & 0.5 (for victim entropy)\\
 Batch Size&   256 \\
 Dataset Size& $10^5$ \\
 Number of Epochs & $10^4$\\
 Gradient Update every... & 1 iteration\\
 Update Target Value Network every... & 100 iterations\\
 Evaluate Progress every... & 500 iterations\\
 Number of Actions Observed for Empirical Victim Entropy... & $10^6$ actions\\

 \hline
\end{tabular}
\caption{Batch Deep Q-Learning Training Hyperparameters}
    \label{tab:batchhyper}
\end{table}

\begin{table}[h]
    \centering
\begin{tabular}{ |p{3.5cm}||p{3.5cm}|}
 \hline
 \multicolumn{2}{|c|}{} \\
 \hline
 Optimizer & Adam\\
 Learning Rate & $10^{-5}$ (for actor)\\
 & $10^{-4}$ (for critic)\\
 Discount Factor $\gamma$   & 0.95\\
 Batch Size&   256 \\
 Replay Buffer Capacity & $5 * 10^5$ \\
 Evaluate Progress every... & 10 games\\

 \hline
\end{tabular}
\caption{MADDPG Training Hyperparameters}
    \label{tab:maddpghyper}
\end{table}

\begin{algorithm}[h]
Initialize dataset $\mathcal{D}$\\
Initialize hashtable $\mathcal{T}$ where states $s\in \mathcal{S}$ are keys, and values are hashtables $\mathcal{T}^{\mathcal{A}}$\\
Initialize each hashtable $\mathcal{T}^{\mathcal{A}}$, where each key is the set of legal actions at state $s$, with corresponding initial value 0\\
Initialize action-value function $Q$ with random weights\\
$\triangleright$ \emph{Exploration Phase}\\
Use Deep-Q Learning to learn a value function $V_\theta$ with reward function $r(s) = H_{0.5}(\pi_\theta(s))$ where $\pi_\theta$ is the softmax of the q-value function $Q_\theta$.\\
$\triangleright$ \emph{Rollout Phase}\\
Rollout policy $\pi_\theta$ against the victim $\nu$. Record $M$ number of trajectories in $\mathcal{D}$. Record $K$ number of victim actions for the victim's empirical entropy in $\mathcal{T}$ by counting each action at each state.\\
$\triangleright$ \emph{Planning Phase}\\
Let $\hat{\pi}(s)$ is the empirical distribution of the victim actions at state $s$ collected in $\mathcal{T}$. Use batch version of Deep-Q Learning on dataset $\mathcal{D}$ using the reward function $r(s) = H_{0.5}(\hat{\pi}(s))$,  or $-1$ if $s$ was not observed during the rollout phase. \\

\caption{Reward-Free RL Algorithm Using R{\'e}nyi Entropy and Empirical Victim Entropy}
\end{algorithm}

\begin{algorithm}[H]
Initialize replay memory $\mathcal{D}$ to capacity $N$\\
Initialize action-value function $Q$ with random weights\\
\For{episode = 1, M}{
Initialise state $s_1$\\
    \For{t = 1, T}{
    Observe new victim value $V_\nu(s_{t})$\\
    With probability $\epsilon$ select a random action $a_t$\\
    Otherwise select $a_t = \argmax_a Q (s_t, a; \theta)$\\
    Execute action $a_t$ in emulator and observe new state $s_{t+1}$ and new victim value $V_\nu(s_{t+1})$\\
    \[
    v_t:=
    \begin{cases}
    -V_\nu(s_{t+1}) & \text{if learning antagonistic value}\\
    Metric(s_t) & \text{if learning to maximize environment-specific metric}\\
    H_{0.5}(softmax(Q_\nu(s_t))) & \text{if learning victim entropy}
    \end{cases}
    \]
    Store transition $(s_t, a_t, v_t, s_{t+1})$ in $\mathcal{D}$\\
    Sample random minibatch of transitions $(s_j , a_j , v_j , s_{j+1})$ from $\mathcal{D}$\\
    \[
    y_j:=
    \begin{cases}
    v_j & \text{for terminal } s_{j+1}, \\
    v_j + \gamma \max_{a'}Q(s_{j+1}, a'; \theta) & \text{for non-terminal } s_{j+1}.
    \end{cases}
    \]
    Perform a gradient descent step on $(y_j - Q(s_{j+1}, a_j, \theta))^2$
    }
}
\caption{Antagonistic Deep Q-Learning}
\end{algorithm}

\newpage

\begin{algorithm}[H]
Initialize replay memory $\mathcal{D}$ to capacity $N$\\
Set agent 1 to be the antagonistic agent\\
Set agent 2 to be the corresponding victim agent with actor policy $\boldsymbol\mu_2$ and actor value function $V_2$\\
\For{episode = 1, T}{
Initialise a random process $\mathcal{N}$ for action exploration\\
Receive initial state $\textbf{x}$\\
    \For{t = 1, max-episode-length}{
    for each agent $i$, select action $a_i = \boldsymbol\mu_i(o_i) + \mathcal{N}_t$ w.r.t the current policy and exploration\\
    Execute actions $a = (a_1,\dots,a_N)$ and observe next state $\textbf{x}'$ and rewards $r_2,\dots, r_N$\\
    Define\\
    \[
    r_1:=
    \begin{cases}
    -V_2(o'_1) & \text{if learning antagonistic value and $o'_1$ is the observation of agent 1 at $\textbf{x}'$}\\
    Metric(o_1) & \text{if learning to maximize environment-specific metric}\\
    -\sum_{i} \boldsymbol\mu_2(a_i | o_1) \log \boldsymbol\mu_2(a_i | o_1) & \text{if learning victim entropy}
    \end{cases}
    \]
    Define $r = (r_1, \dots, r_N)$.\\
    Store transition $(\textbf{x}, a, r, \textbf{x}')$ in $\mathcal{D}$\\
    $\textbf{x} \leftarrow \textbf{x}'$\\
    \For{\text{agent} i = 1, N}{
        Sample random minibatch of of $S$ samples $(\textbf{x}^j, a^j, r^j, \textbf{x}'^j)$ from $\mathcal{D}$\\
        Set $y^j = r^j_i + \gamma Q^{\boldsymbol\mu'}_i(\textbf{x}'^j, a'_1,\dots, a'_N)|_{a'_k=\boldsymbol\mu'_k(o^j_k)}$\\
        Update critic by minimizing the loss $\mathcal{L}(\theta_i) = \dfrac{1}{S}\sum_j (y^j - Q^{\boldsymbol\mu}_i(\textbf{x}, a^j_1,\dots, a^j_N))^2$\\
        Update actor using the sampled policy gradient: $\nabla_{\theta_i}J \approx \dfrac{1}{S}\sum_j\nabla_{\theta_i}\boldsymbol\mu_i(o^j_i)\nabla_{a_i}Q^{\boldsymbol\mu}_i(\textbf{x}^j, a^j_1, \dots, a_i,\dots, a^j_N) |_{a_i=\boldsymbol\mu_i(o^j_i)}$
        }
    Update target network parameters for each agent $i$: $\theta'_i \leftarrow \tau\theta_i + (1 - \tau)\theta'_i$
    }
}
\caption{Multi-agent Deep Deterministic Policy Gradient for $N$ agents (including 1 antagonist)}
\end{algorithm}

\newpage

\section{Figures and Results}


\begin{figure}[!htb]
\centering
\includegraphics[scale=0.2]{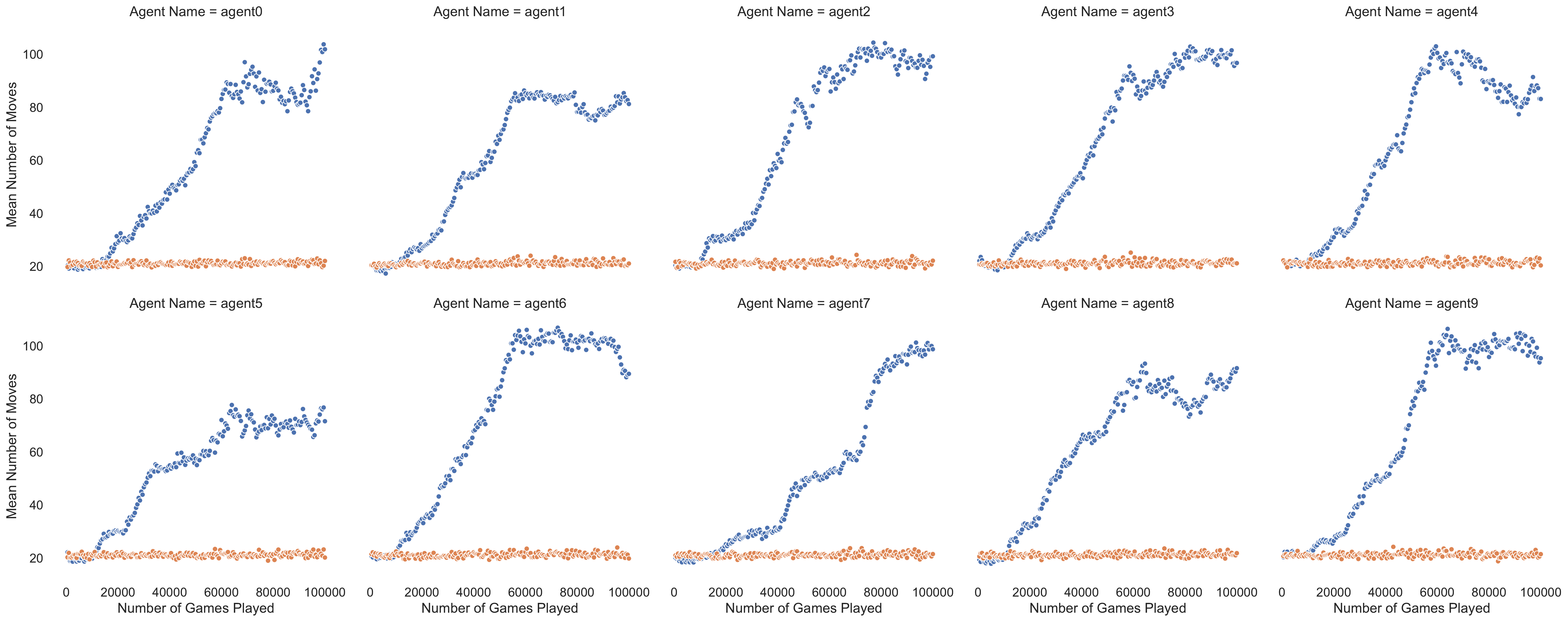}
  \caption{Each plot shows each Breakthrough agent's number of moves of each type of agent with the same random seed. \\
  All normal DQNs (in orange) show no increase in number of moves throughout the training process. All antagonistic value DQNs (in blue), move-maximizing DQNs (in green), and victim entropy DQNs (in red) show an increase in the number of moves.}
\end{figure}

\begin{figure}[!htb]
\centering
\includegraphics[scale=0.2]{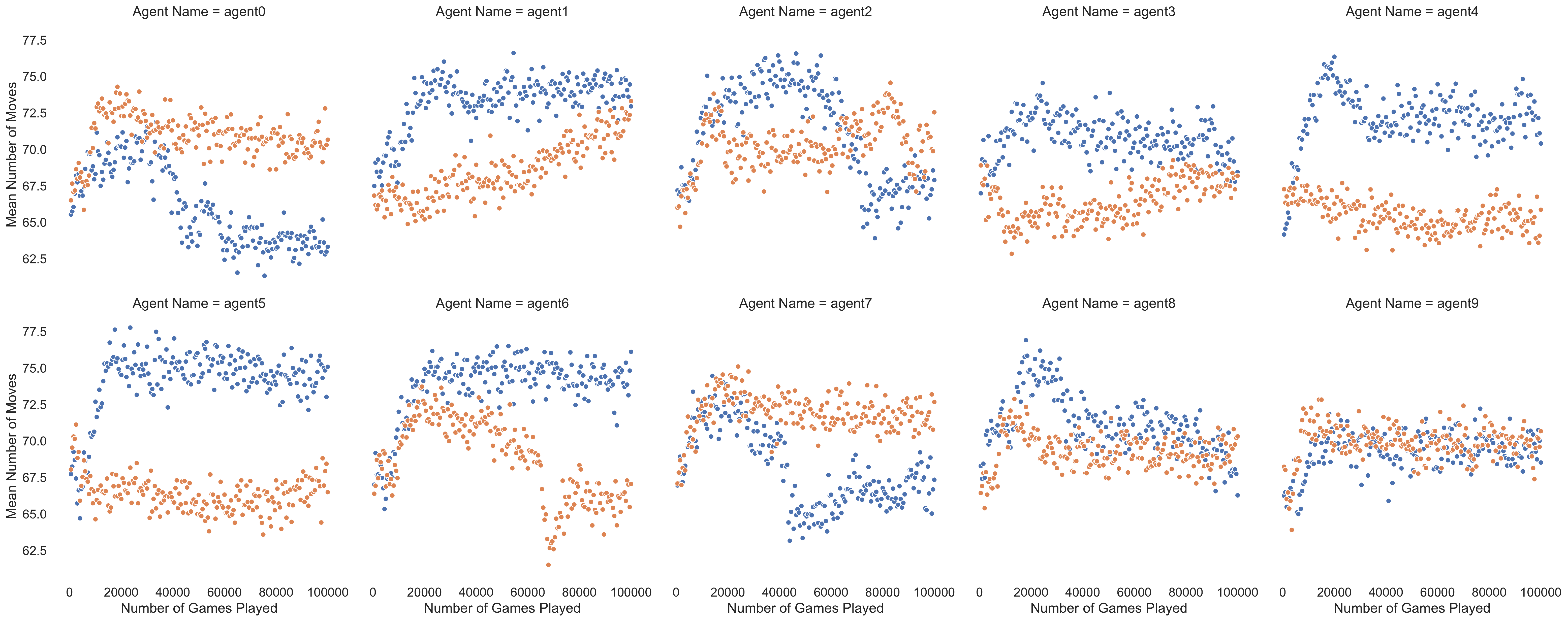}
  \caption{Havannah number of moves of each type of agent with the same random seed. \\
  Most of the normal DQNs (in orange) show no increase in number of moves throughout the training process. Most antagonistic value DQNs (in blue) and DQNs that learn to maximize number of moves (in green) show an increase in the number of moves. The victim entropy DQNs (in red) all show a decline in the number of moves.}
\end{figure}

\end{document}